# Initial Study into Application of Feature Density and Linguistically-backed Embeddings to Improve Machine Learning-based Cyberbullying Detection

Juuso Eronen[1], Michal Ptaszynski[1], Fumito Masui[1], Gniewosz Leliwa[2], Michal Wroczynski[2], Mateusz Piech[3] and Aleksander Smywinski-Pohl[3]

[1]Kitami Institute of Technology
[2]Samurai Labs/Fido Voice
[3]AGH University of Science and Technology

eronen.juuso@gmail.com, {ptaszynski, f-masui}@cs.kitami-it.ac.jp, {gniewosz.leliwa, michal.wroczynski}@fido.ai, {mpiech, apohllo}@agh.edu.pl

## Abstract

In this research, we study the change in the performance of machine learning (ML) classifiers when various linguistic preprocessing methods of a dataset were used, with the specific focus on linguistically-backed embeddings in Convolutional Neural Networks (CNN). Moreover, we study the concept of Feature Density and confirm its potential to comparatively predict the performance of ML classifiers, including CNN. The research was conducted on a Formspring dataset provided in a Kaggle competition on automatic cyberbullying detection. The dataset was re-annotated by objective experts (psychologists), as the importance of professional annotation in cyberbullying research has been indicated multiple times. The study confirmed the effectiveness of Neural Networks in cyberbullying detection and the correlation between classifier performance and Feature Density while also proposing a new approach of training various linguistically-backed embeddings for Convolutional Neural Networks.

## 1 Introduction

The popularity Artificial Intelligence (AI) and Machine Learning (ML) has gained throughout recent years has also lead to new challenges. Training classifiers on large datasets is both time consuming and computationally intensive while leaving behind a noticeable carbon footprint [Strubell et al., 2019]. To move towards greener AI [Schwartz et al., 2019], we need to take a look at the core methods of ML and find potential points of improvement. In order to save computational resources, it would be useful to roughly predict the performance of a classifier before the training.

In the recent years Convolutional Neural Networks, in addition to their original use in image recognition and computer vision [LeCun. et al., 1989], have shown their effectiveness in Natural Language Processing (NLP) and text classification [Collobert et al., 2011]. Numerical word representation, or vectorization, is one of the key concepts in NLP and text classification and is most commonly performed by using traditional methods like transforming the documents into a matrix of token-vectors (Bag-of-Words, BoW) usually created using a weight calculation scheme like TF-IDF (term frequency with inverse document frequency). However, these will not take the semantic or syntactic relationships into account.

With Neural Word Embeddings it is possible to capture deeper relations between words. Instead using linear algebra, like TF-IDF, the embeddings are learned during training of a Neural Network model [Bengio et al., 2003]. This allows capturing of similarities and relations between words. Word embeddings trained as a part of a Neural Network model help keeping the the semantic and syntactic information of words by grouping words representing similar contexts closer together in the produced vector space.

Feature Density (FD) was initially introduced by [Ptaszynski et al., 2017] to help estimate classifier performance by looking at the amount of unique features compared to the total number of features in a dataset. They discovered that some classifiers (CNN) benefited from higher FD while others performed better with lower FD. However, their research was based only on a dataset in Japanese. In this research we aim at verifying the usefulness of FD also for English.

The dataset used in this research, based on Formspring.me, was originally annotated using Amazon Mechanical Turk and applied in a Kaggle competition on automatic cyberbullying detection. We re-annotated the dataset by experts (psychologists) as a part of this research to ensure quality annotations, as in such research it is crucial that the dataset is annotated by professionals either experienced in Internet Patrol activities (patrolling the Internet in the search for harmful contents), or having a sufficient expert psychological knowledge. The dataset is comparatively large in size with over twelve thousand samples in total. The class distribution also roughly represents that of a real life situation, with around 7% of entries being labeled for cyberbullying.

In this research we assess the performance of various ML classifiers using different linguistically-backed embeddings trained as a part of a neural network for CNNs and by applying TF-IDF on a BoW for other classifiers. Also, we study the effect of FD and verify its potential in comparatively estimating the performance of classifiers. The preprocessing methods applied here were the same as in previous research [Ptaszynski et al., 2017] for better comparability, this time with an English dataset. This also allows us to check how the linguistic and cultural differences affect the classifier performance when same classification methods are used.

The paper is organized in the following way. Firstly, we present the previous research on Linguistically-backed em-beddings and Feature Density. Further, we present the dataset used in this research, and explain the evaluation settings, followed by the analysis of experiment results and discussion.

## 2 Research Background

### 2.1 Feature Density

A dataset is the more generalized, the fewer number of frequently appearing unique features it contains. [Ptaszynski *et al.*, 2017] defined Feature Density (FD) by applying the notion of Lexical Density [Ure, 1971] from linguistics. It is a score representing an estimated measure of content per lexical units for a given corpus, calculated as the number of all unique words divided by the number of all words in the corpus. The term Feature Density comes from the fact that a variety of different features is used, not only words. After calculating FD for all applied dataset preprocessing methods we calculated Pearson's correlation coefficient ($\rho$-value) between dataset generalization (FD) and classifier results (F-scores).

If FD correlates with classifier performance (either positively or negatively) it could be useful in roughly estimating the performance in on datasets with various FD. For example, [Ptaszynski *et al.*, 2017] showed that CNN benefits from higher FD while other classifiers' score was higher using datasets with lower FD. This suggests that the performance of CNNs could be improved by increasing the FD of the applied dataset while other classifiers could see better results with lower FD [Ptaszynski *et al.*, 2017].

Increasing or decreasing the base FD of a dataset could have uses in finding an optimal amount of information that could be used to improve classifier performance. The Feature Density of a dataset can be manipulated by the use of different kinds of preprocessing methods that either add information or generalize the dataset. For example, FD can be increased by adding parts-of-speech information to tokens. On the other hand, FD can be reduced by lemmatizing.

### 2.2 Linguistically-backed Word Embeddings

Almost without exception, the word embeddings are learned from pure tokens (words) or lemmas (unconjugated forms of words). This is also the case with the most well-known semi-supervised neural embedding frameworks like Word2vec [Mikolov *et al.*, 2013] and GloVe [Pennington *et al.*, 2014].

To investigate the potential of capturing deeper relations between lexical items and structures we propose a novel method of adding linguistic information to the pure tokens or lemmas in order to preserve the morphological, syntactic and other types of information. Examples would include combining the tokens with their respective parts-of-speech or dependency information. These combinations would then be used to train the word embeddings. The preprocessing methods are described in-depth in section 5.1. To the best of our knowledge, embeddings backed with linguistic information have not yet been researched extensively, with only a handful of related work [Levy and Goldberg, 2014; Komninos and Manandhar, 2016; Cotterell and Schütze, 2019].

## 3 Cyberbullying

### 3.1 Description of the Problem

Cyberbullying (CB) is a phenomenon observed in many Social Networking Services (SNS), where users (often young), share their thoughts regarding their personal interests, life and problems. Some of the users engage in activities such as harassment, threat, intimidation and mocking in order to make others uncomfortable, undermine their self-esteem and to discourage them from engaging in the online conversation. Anonymity on the Internet is one of the factors that make this activity attractive for bullies since they rarely face social ostracism or other consequences of their improper behavior. The problem has been growing since the outset of SNS.

As such CB is an important problem, which has its roots in technology. However, technology can also help reduce or eradicate it. Development of Natural Language Processing (NLP) algorithms aimed at the detection of CB is one of the ways to achieve that goal. Such algorithms require properly annotated data to assure their performance. Moreover, the recently most popular Machine Learning (ML) algorithms, Deep Neural Networks (DNN) in particular, require large annotated corpora in order to obtain high quality classification results. Unfortunately, even though there is a number of publicly available datasets for general hate speech detection, only a few of them concern the specific topic of CB detection:

- Kaggle Formspring data for Cyberbullying Detection [1]
- MySpace Group Data Labeled for Cyberbullying [2]
- CREEP EIT project Whatsapp Cyberbullying dataset [3]
- PolEval 2019 Cyberbullying dataset [4]

### 3.2 Research on Cyberbullying Detection

The first recorded study on cyberbullying detection discovered that vulgar words are the most distinctive features of cyberbullying [Ptaszynski *et al.*, 2010]. These were used to train an Support Vector Machine (SVM) classifier. Other research [Dinakar *et al.*, 2012; Sood *et al.*, 2012] also favored SVMs as the most effective classifiers. This was succeeded by a method using seed words to calculate SO-PMI-IR score and maximize the relevance of categories to rank Internet entries according to their harmfulness [Nitta *et al.*, 2013]. This method was later outperformed by a sentence pattern extraction solution using a brute force search algorithm that both improved performance and reduced human effort [Ptaszynski *et al.*, 2015]. The latest improvement in cyberbullying detection is a CNN based approach which successfully outperformed all previous methods [Ptaszynski *et al.*, 2017].

## 4 Applied Dataset

The Kaggle Formspring dataset applied in this research was originally annotated using the Mechanical Turk service [Reynolds *et al.*, 2011]. The methodology behind the annotation process was simple. Namely, a post was marked as containing cyberbullying if two of the annotators indicated that it does. Moreover, the annotators did not have any training towards the detection of cyberbullying. As a result, the quality of the annotation can be doubted. Since in the research the quality of the dataset and its annotation are crucial for the task, we have decided to use the same dataset, but provided a new annotation, obtained by a well designed process.

---

[1] https://www.kaggle.com/swetaagrawal/formspring-data-for-cyberbullying-detection
[2] http://www.chatcoder.com/Data/BayzickBullyingData.rar
[3] https://github.com/dhfbk/WhatsApp-Dataset
[4] http://2019.poleval.pl/index.php/tasks/task6

Formspring Dataset for Cyberbullying Detection was chosen for the following reasons.
- It is of comparatively large size: 12,772 samples.
- The fact that it is a fairly well-known dataset initially released in October, 2016, last updated in January, 2017.
- The fact that the number of sentences with bullying reflects the real-world proportions of cyberbullying and non-cyberbullying content (more than 84% of samples were labeled as "no cyberbullying").
- The option of anonymity that encourages cyberbullying and other harmful behaviors (Formspring allowed users to post questions anonymously to any other user's page).
- The controversies around Formspring related to harassment and cyberbullying that eventually led to suicides in 2011 and shutdown of the service in 2013.

### 4.1 Previous Annotation
A preliminary analysis of the annotation quality demonstrates numerous shortcomings that put in question its usage in a testing process. Each sample was labeled by three Amazon's Mechanical Turk workers with "yes" and "no" answers for a question if it contains cyberbullying. A post was considered harmful if at least two out of three annotators answered "yes." As a result 802 samples out of 12,772 (6.3%) were classified as "cyberbullying".

In the description of the annotation process there was no information about annotators' competence and there were too many missed cases of cyberbullying as well as many cases of non-bullying content that were incorrectly labeled as cyberbullying. An analysis of the annotated samples showed that at least 2.5% of posts classified as "non-cyberbullying" should be labeled as "cyberbullying" due to their possible harmful impact. This is a noticeable amount compared to the percentage of the samples labeled as "cyberbullying". Similarly, about 15-20% of the samples labeled as "cyberbullying" could be labeled as "non-cyberbullying" due to an infinitesimal harmfulness or even obvious annotators' errors. Therefore, we decided to re-annotate the whole Formspring dataset.

### 4.2 Improved Annotation
The task of Cyberbullying Detection (later abbreviated to CB-D) is specific in the sense that it requires highly trained data annotators with sufficient background for high quality annotations. Differently to well known tasks, such as traditional sentiment analysis, annotators employed in CB-D should either be experienced in Internet Patrol activities (patrolling the Internet in the search for harmful contents), or should have a sufficient professional knowledge in psychology, psychiatry, or related fields.

We made an open call for data annotators within graduate students of psychology, with a condition of at least near-native English proficiency (language of initial data samples). Sixteen (16) initial candidates responded to our call. The candidates were given an initial test to eliminate low performance annotators. In the initial test the candidates were given 30 random samples to annotate with already prepared gold standard answers. The top eight candidates were retained.

### 4.3 Dataset Properties
Table 1 reports some key statistics of the new annotation of the dataset. Statistics described as harmful and non-harmful refer to the final version of the new annotations. The dataset contains approximately 300 thousand of tokens. There were no visible differences in length between the posted questions and answers (approx. 12 words). On the other hand, the harmful samples were usually slghtly shorter than the non-harmful samples (approx. 23 vs. 25 words). The number of harmful samples was small, amounting to 7%, which roughly reflects the amount of profanity on SNS.

Table 1: Statistics of the dataset after improved annotation.

| Element type | Value |
| --- | --- |
| Number of samples | 12,772 |
| Number of harmful samples | 913 |
| Number of non-harmful samples | 11,859 |
| Number of all tokens | 301,198 |
| Number of unique tokens | 18,394 |
| Avg. length (chars) of a single post (Q + A) | 12.1 |
| Avg. length (words) of a single post (Q + A) | 23.6 |
| Avg. length (chars) of a single question | 61.6 |
| Avg. length (words) of a single question | 12 |
| Avg. length (chars) of a single answer | 58.5 |
| Avg. length (words) of a single answer | 11.5 |
| Avg. length (chars) of a harmful post | 12.1 |
| Avg. length (words) of a harmful post | 22.9 |
| Avg. length (chars) of a non-harmful post | 13.9 |
| Avg. length (words) of a non-harmful post | 24.7 |

Apart from the linguistic differences, the most striking difference in the dataset properties when compared to the Japanese dataset from the previous studies [Ptaszynski et al., 2015; Ptaszynski et al., 2017; Ptaszynski and Masui, 2018] was the ratio of harmful samples to non-harmful. The English dataset has only 7% of harmful samples compared to the near 50% in the Japanese dataset. This needs to be taken into account when evaluating the results of this study.

### 4.4 Comparison with Previous Annotation
The original annotation contained 802 samples out of 12,772 (6.3%) labeled as cyberbullying/harmful when at least 2 out of 3 layperson annotators agreed [Reynolds et al., 2011]. The new annotation contains 913 samples out of 12,772 (7.1%) labeled as harmful by expert annotators.

There are 392 samples that were labeled as non-harmful in the original annotation and as harmful in the new annotation. There are 281 samples that were labeled as harmful in the original annotation and as non-harmful in the new annotation.

## 5 Proposed Methods
### 5.1 Dataset Preprocessing
In order to train the linguistically-backed embeddings, we first preprocessed the dataset in various ways, similarly to [Ptaszynski et al., 2017]. This was done for three reasons. Firstly, to see how traditional classifiers managed the data from similar domain (cyberbullying), but in different language. Secondly, to later verify the correlation between the classification results and Feature Density (FD). Finally, to verify the performance of various versions of the proposed linguistically-backed embeddings. Also, since we aimed at making the proposed approach fully systematic and automatic, we did not rely on any hand-made features, such as offensive word lexicons, etc, used in previous research [Ptaszynski et al., 2010]. The preprocessing was done using spaCy NLP toolkit (https://spacy.io/).

- **Tokenization:** includes words, punctuation marks, etc. separated by spaces (later: TOK).
- **Lemmatization:** like the above but with generic (dictionary) forms of words ("lemmas") (later: LEM).
- **Parts of speech:** Words are represented as their representative parts of speech (later: POS).
- **Tokens with POS:** words with POS information is included in each feature (later: TOK+POS).
- **Lemmas with POS:** like the above but with lemmas (later: LEM+POS).
- **Tokens with Named Entity Recognition:** words encoded together with with information on what named entities (private name of a person, organization, numericals, etc.) appear in the sentence (later: TOK+NER).
- **Lemmas with NER:** like the above but with lemmas (later: LEM+NER).
- **Dependency structure:** noun- and verb-phrases with syntactic relations between them (later: DEP).
- **Chunking:** like above but without dependency relations ("chunks", later: CHNK).
- **Chunking with NER:** information on named entities is encoded in chunks (later: CHNK+NER).
- **Dependency structure with Named Entities:** both dependency relations and named entities are included in each feature (later: DEP+NER).

### 5.2 Feature Extraction

For the non-CNN classifiers, from each of the eleven processed dataset versions, a Bag-of-Words language model was generated, producing a separate model for each of the datasets (Bag-of-Words, Bag-of-Lemmas, Bag-of-POS, etc.). The language models generated from the entries of the dataset were used later in the input layer of classification. We also applied a traditional weight calculation scheme, namely term frequency with inverse document frequency $tf * idf$, where term frequency $tf(t, d)$ refers to raw frequency (number of times a term $t$ (word, token) occurs in a document $d$), and inverse document frequency $idf(t, D)$ is the logarithm of the total number of documents $|D|$ in the corpus divided by the number of documents containing the term $n_t$. Finally, $tf*idf$ refers to term frequency multiplied by inverse document frequency as in equation (1).

$$idf(t, D) = log(\frac{|D|}{n_t}) \qquad (1)$$

When training a Convolutional Neural Network model, the embeddings were trained as a part of the network for all of the described datasets. Similarly to other classifiers, we trained a separate model for each of the 68 datasets (Word/token Embeddings, Lemmas Embeddings, POS Embeddings, Chunks Embeddings, etc.). The embeddings were trained as part of the network using Keras' embedding layer with random initial weights.

### 5.3 Classification

**SVM** or Support-vector machines [Cortes and Vapnik, 1995] are a set of classifiers well established in AI and NLP. They represent data, belonging to specified categories, as points in space (vectors), and find an optimal hyperplane to separate the examples from each category. SVM has had much success in previous cyberbullying research [Ptaszynski et al., 2010]. We used the linear SVM function, which finds the maximum-margin hyperplane dividing the samples, as it had the best performance out of a number of SVM kernel functions in previous research [Ptaszynski et al., 2017].

**NaïveBayes** (NB) classifier is a supervised learning algorithm applying Bayes' theorem with the assumption of a strong (naïve) independence between pairs of features, traditionally used as a baseline in text classification tasks. It is known for working well with smaller datasets and it is fast to train compared to other classifiers e.g. Random Forest.

**kNN** or the k-Nearest Neighbors classifier takes as input k-closest training samples with assigned classes and classifies input sample by a majority vote. It is often applied as a baseline, next to Naïve Bayes. The classifier is fast and simple to train but is very susceptible to outliers and overfitting. Here, we used k=1 setting in which the input sample is assigned to the class of the first nearest neighbor.

**Random Forest** (RF) in training phase creates multiple decision trees to output the optimal class (mode of classes) in classification phase [Breiman, 2001]. An improvement of RF to standard decision trees is the ability to correct overfitting to the training set common in decision trees [Hastie et al., 2013]. In practice, Random Forest starts by taking a random bootstrap sample with replacement from the dataset and then selects a random subset of features to reduce the dimensionality of the sample. Next, an unpruned decision tree is trained on this bootstrap sample. This process is repeated for the desired ensemble size. The predicted value of an unknown instance is obtained by taking a majority vote [Breiman, 2001].

**Logistic Regression** (LR) is a statistical model that calculates class probabilities using a logistic function (sigmoid) instead of a straight line or hyperplane. It assigns a probability value between [0,1] for each data point which is used in assigning it to a class. Logistic regression models are usually fit using Maximum Likelihood Estimation [Hastie et al., 2013].

**MLP** (Multilayer Perceptron) is a type of feed-forward artificial neural network consisting of an input layer, an output layer and one or more hidden layers. In this experiment MLP refers to a network using regular dense layers. We applied an MLP implementation with Rectified Linear Units (ReLU) as a neuron activation function [Hinton et al., 2012] and one hidden layer with dropout regularization which reduces overfitting and improves generalization by randomly dropping out some of the hidden units during training [Hinton et al., 2012].

**CNN** or Convolutional Neural Networks are a type of feed-forward artificial neural network utilizing convolutional and pooling layers. Although originally CNN were designed for image recognition, their performance has been confirmed in many tasks, including NLP [Collobert and Weston, 2008] and sentence classification [Kim, 2014]. We applied a CNN implementation with Rectified Linear Units (ReLU) as a neuron activation function, and max pooling [Scherer et al., 2010], which applies a max filter to non-overlying sub-parts of the input to reduce dimensionality and in effect correct overfitting. We also applied dropout regularization on penultimate layer. An embedding layer with random initial weights is used to train the word embeddings. We applied two versions of CNN. First, with one hidden convolutional layer containing 128 units was applied as a baseline. Second, the final proposed method consisted of two hidden convolutional layers, containing 128 feature maps each, with 4x4 size of patch and 2x2 max-pooling, and Adaptive Moment Estimation (Adam), a variant of Stochastic Gradient Descent [LeCun et al., 2012].

# 6 Experiments

## 6.1 Experiment Setup

The class distribution was kept as-is to keep the experiment as close to a real-life situation as possible. In the case of SVM and Neural Networks, the class balance issue was dealt with by setting the class weights in order to punish the misclassification of the minority class more. The preprocessed dataset provides eleven separate datasets and the experiment was performed once for each preprocessing type. Each of the classifiers (sect. 5.3) were tested on each version of the dataset in a 5-fold cross validation procedure. The results were calculated using standard Precision (Prec), Recall (Rec), balanced F-score (F1), Accuracy (Acc) and Area Under Receiver Operating Characteristic Curve (AUC). As for the winning condition, we looked at which classifier achieved the highest F-score. We used Feature Density to evaluate the effects of different preprocessing types on the results to verify its potential in estimating classifier performance [Ptaszynski et al., 2017].

## 6.2 Results and Discussion

**Classifier Results**

All results were summarized in Table 3. The results of the baselines (kNN, Naïve Bayes) were the lowest. Even though these classifiers can typically reach high scores in text classification, they were not able to grasp the noisy language of cyberbullying and had the worst scores across the board.

Surprisingly, Random Forest also scored very low. For some datasets it had even lower scores than kNN and Naïve Bayes. This differs from the results of previous research [Ptaszynski et al., 2017] where RF had better than average scores for the Japanese cyberbullying dataset. RF is also very time inefficient to train compared to SVM.

In many previous research on CB detection, SVM were the most commonly used with various success. Also here, SVM had generally high scores and close to the highest scores achieved by LR and CNN. SVM had the best score on TOK+NER (F1=.782). Although not scoring the highest, considering the ratio of training time to the results, SVM can still be considered one of the most efficient classifiers.

The highest score was achieved by LR (F1=.798), slightly higher than SVM in almost all of the datasets except those with highest FD 2. Despite being a very simple classifier, LR's performance was surprisingly high as it overtook the proposed CNN in most of the datasets, although most often with only a small margin. Considering its overall performance and time-efficiency, LR can be considered the best classifier of the experiment and will be used in the future.

As for Neural Network-based classifiers, MLP trained with two hidden layers on the datasets reached average scores, generally slightly lower than SVM. Its scores did not leave much to note and due to it being time-inefficient to train, we will not consider MLP in our future experiments.

The proposed method, CNNs trained on various linguistically-backed embeddings, also scored high across the board. Unlike in previous research [Ptaszynski et al., 2017], however, there was no major difference between one and two layered networks. The scores of CNNs were generally only very slightly lower than LR. Interestingly, CNNs were the only classifiers that also achieved reasonably high scores on the highest FD datasets (namely DEP and DEP+NER), which is in line with previous studies [Ptaszynski et al., 2017]. However, among all embeddings the most typical plain word embeddings based on tokens or lemmas achieved the highest scores with the CNNs for English.

Table 2: Left: Feature Density (FD) of each preprocessing type. Top Right: Pearson Correlation Coefficient ($\rho$-value) of FD with all classifier results, with statistical significance (2-sided p-value), with normalized data for non-NeuralNet classifiers. Bottom Right: Student's t-test statistical significance (2-sided p-value) between top 3 best performing classifiers.

| Preprocessing Type | Unique 1-grams | All 1-grams | Feature Density |
|---|---|---|---|
| DEP | 149360 | 309592 | .4824 |
| DEP+NER | 147560 | 309592 | .4766 |
| CHNK | 39017 | 309592 | .1260 |
| CHNK+NER | 33612 | 309592 | .1085 |
| TOK+POS | 31474 | 367139 | .0857 |
| LEM+POS | 26598 | 367180 | .0724 |
| TOK | 25785 | 367139 | .0702 |
| LEM | 21766 | 367180 | .0592 |
| TOK+NER | 21627 | 367102 | .0589 |
| LEM+NER | 17696 | 367142 | .0482 |
| POS | 19 | 357604 | .0001 |

| Classifier | $\rho$-value | 2-sided p-value |
|---|---|---|
| CNN-2L | .1558 | .6474 |
| CNN-1L | .0837 | .8067 |
| MLP | -.2126 | .5301 |
| NB | -.2617 | .4368 |
| LR | -.2991 | .3716 |
| SVM | -.499 | .1181 |
| RF | -.6737 | .0230 |
| kNN | -.7671 | .0058 |

| Classifiers | 2-sided p-value |
|---|---|
| CNN-2L&SVM | .9162 |
| CNN-2L&LR | .6189 |
| SVM&LR | .6831 |

As the dataset was not balanced, we also tried oversampling the minority class using Synthetic Minority Oversampling Technique (SMOTE) [Chawla et al., 2002] and ran the experiments again for classifiers other than MLP and CNN. The results were represented in Table 4. Oversampling the data showed some performance changes, although the three best classifiers remained the same. Interestingly, LR's performance degraded and it was overtaken by both CNN and SVM. This would mean that if one considers both real-life and laboratory conditions, SVM could be considered the best classifier over LR. Another change was the noticeable performance boost of RF. The bottom right part of Table 2 shows that there is no statistical significance between the results of the top three best performing classifiers so the rank order of these classifiers could vary when compared on different data.

From the results it can be seen that most of the classifiers scored highest on pure tokens or lemmas. This is different from previous research performed for Japanese, where the performance of tokens and lemmas were always improved by named entity recognition. This could be explained by the way cyberbullying differs between cultures; in Japan it is most often realised as exposing a person's sensitive information such as their name and address (also known as "doxing"), whereas in English it is often realised as a personal attack usually with an accumulation of profanities and toxic vocabulary. This difference is also supported by the official governmental definition of cyberbullying in Japan, which states that CB often contains revealing of private information [MEXT, 2008].

**Effect of Feature Density and Linguistic Embeddings**

Next, we analyzed the correlation of data preprocessing with Feature Density. The results were represented in Table 2. Similarly to previous research [Ptaszynski et al., 2017], both kNN and RF correlated strongly negatively with FD, with the result also being statistically significant. The rest of the non-CNN classifiers had a weak negative correlation with FD, although these results did not show statistical significance. The CNNs had a completely different degree of correlation compared to previous study, both of them having a weak positive correlation with FD. This could be because of the difference in the applied CNN settings. Previous research ap-plied simple BoW, whereas the proposed method used various linguistically-backed embeddings trained as a part of the network instead of a BoW approach. With the linguistic embeddings, increasing FD did not seem to increase the classifier performance, at least for the English dataset.

Table 3: Scores of all applied classifiers (Averaged for positive & negative prediction calculated separately; best classifier for each dataset in **bold**; best dataset generalization for each underlined).

| | | TOK | LEM | LEM+NER | POS | TOK+NER | TOK+POS | LEM+POS | CHNK | CHNK+NER | DEP | DEP+NER |
|---|---|---|---|---|---|---|---|---|---|---|---|---|
| RF | Acc | .942 | .943 | .94 | .928 | .942 | .938 | .941 | .935 | .934 | .93 | .931 |
| | Prec | .899 | .887 | .844 | .549 | .91 | .87 | .906 | .886 | .843 | .965 | .903 |
| | Rec | .611 | .602 | .595 | .502 | .595 | .555 | .59 | .54 | .541 | .516 | .519 |
| | F1 | .663 | .652 | .638 | .487 | .643 | .581 | .635 | .559 | .558 | .514 | .519 |
| | AUC | .612 | .62 | .6 | .507 | .606 | .572 | .599 | .555 | .549 | .515 | .516 |
| SVM | Acc | .949 | .937 | .924 | .68 | .936 | .943 | .933 | .939 | .931 | .933 | .935 |
| | Prec | .796 | .751 | .719 | .541 | .769 | .789 | .757 | .768 | .754 | .826 | .793 |
| | Rec | .796 | .795 | .765 | .653 | .797 | .757 | .761 | .704 | .687 | .566 | .558 |
| | F1 | **.796** | .771 | .739 | .508 | **.782** | .771 | .759 | .73 | .714 | .598 | .585 |
| | AUC | .816 | .801 | .775 | .635 | .792 | .755 | .749 | .704 | .676 | .565 | .57 |
| NB | Acc | .932 | .932 | .929 | .929 | .929 | .929 | .929 | .929 | .929 | .929 | .929 |
| | Prec | .966 | .895 | .798 | .464 | .715 | .964 | .965 | .465 | .964 | .464 | .464 |
| | Rec | .519 | .516 | .511 | .5 | .505 | .508 | .505 | .5 | .503 | .5 | .5 |
| | F1 | .52 | .514 | .503 | .482 | .493 | .498 | .492 | .482 | .487 | .482 | .482 |
| | AUC | .523 | .526 | .508 | .5 | .508 | .504 | .504 | .503 | .501 | .5 | .5 |
| kNN | Acc | .933 | .929 | .928 | .879 | .927 | .924 | .926 | .932 | .929 | .898 | .902 |
| | Prec | .809 | .742 | .679 | .53 | .679 | .717 | .69 | .833 | .676 | .6 | .62 |
| | Rec | .54 | .527 | .557 | .533 | .526 | .662 | .653 | .53 | .534 | .604 | .609 |
| | F1 | .556 | .536 | .576 | .531 | .534 | .685 | .668 | .538 | .543 | .602 | .614 |
| | AUC | .548 | .532 | .553 | .533 | .534 | .657 | .654 | .53 | .532 | .604 | .599 |
| LR | Acc | .942 | .936 | .929 | .788 | .926 | .936 | .934 | .932 | .92 | .924 | .928 |
| | Prec | .781 | .733 | .702 | .518 | .683 | .746 | .73 | .73 | .662 | .653 | .704 |
| | Rec | .816 | .819 | .827 | .77 | .786 | .803 | .784 | .773 | .732 | .587 | .628 |
| | F1 | **.798** | **.773** | **.756** | **.571** | .729 | **.773** | .756 | **.751** | .695 | .601 | .649 |
| | AUC | .69 | .695 | .68 | .549 | .663 | .662 | .658 | .637 | .61 | .538 | .559 |
| MLP | Acc | .931 | .923 | .919 | .787 | .921 | .921 | .921 | .905 | .896 | .871 | .88 |
| | Prec | .72 | .731 | .714 | .531 | .704 | .729 | .726 | .671 | .64 | .599 | .608 |
| | Rec | .79 | .746 | .802 | .377 | .769 | .764 | .767 | .705 | .723 | .662 | .648 |
| | F1 | .75 | .738 | .749 | .44 | .731 | .744 | .744 | .686 | .671 | .621 | .623 |
| | AUC | .905 | .882 | .902 | .627 | .88 | .885 | .883 | .85 | .83 | .775 | .781 |
| CNN-1L | Acc | .932 | .925 | .92 | .386 | .919 | .917 | .924 | .925 | .92 | .903 | .905 |
| | Prec | .762 | .723 | .715 | .518 | .692 | .718 | .729 | .7 | .718 | .66 | .674 |
| | Rec | .814 | .807 | .8 | .791 | .753 | .796 | .815 | .79 | .782 | .788 | .775 |
| | F1 | .785 | .757 | .749 | .451 | .717 | .749 | **.764** | .736 | **.745** | .702 | .709 |
| | AUC | .893 | .885 | .883 | .672 | .878 | .866 | .885 | .878 | .881 | .87 | .846 |
| CNN-2L | Acc | .924 | .916 | .894 | .696 | .913 | .929 | .92 | .923 | .898 | .905 | .915 |
| | Prec | .718 | .74 | .652 | .534 | .698 | .728 | .711 | .718 | .662 | .692 | .686 |
| | Rec | .828 | .802 | .791 | .608 | .806 | .79 | .793 | .793 | .754 | .749 | .766 |
| | F1 | .761 | .764 | .694 | .503 | .737 | .755 | .743 | .747 | .694 | **.713** | **.718** |
| | AUC | .893 | .894 | .879 | .677 | .882 | .89 | .908 | .908 | .871 | .859 | .861 |

Table 4: F1 scores of non-NeuralNet classifiers on data with SMOTE (Original NeuralNet results for reference, averaged for positive & negative prediction calculated separately; best classifier for each dataset in **bold**; best dataset generalization for each underlined).

| | TOK | LEM | LEM+NER | POS | TOK+NER | TOK+POS | LEM+POS | CHNK | CHNK+NER | DEP | DEP+NER |
|---|---|---|---|---|---|---|---|---|---|---|---|
| RF | .72 | .696 | .675 | **.55** | .692 | .645 | .656 | .58 | .571 | .529 | .508 |
| SVM | .779 | **.792** | **.749** | .5 | .727 | **.766** | .749 | .7 | .71 | .592 | .561 |
| NB | .717 | .719 | .688 | .502 | .689 | .717 | .7 | .68 | .656 | .616 | .612 |
| kNN | .598 | .637 | .597 | .534 | .605 | .338 | .36 | .59 | .552 | .245 | .239 |
| LR | .782 | .759 | .722 | .487 | .737 | .761 | .752 | .75 | .677 | .609 | .637 |
| MLP | .75 | .738 | .749 | .44 | .731 | .744 | .744 | .69 | .671 | .621 | .623 |
| CNN-1L | **.785** | .757 | .749 | .451 | .717 | .749 | **.764** | .74 | **.745** | .702 | .709 |
| CNN-2L | .761 | .764 | .694 | .503 | **.737** | .755 | .743 | **.75** | .694 | **.713** | **.718** |

Comparing the results with our previous studies shows that the optimal amount of information the classifier should be trained with is language-dependent. In case of English, generalizing the data by lemmatizing or by applying NER seemed to lower the performance in most cases while with Japanese the results were completely opposite [Ptaszynski et al., 2017]. Also, after increasing FD by including dependency information, the scores were very high with CNN on the Japanese dataset. For English, the results of CNN with high FD dataset were only average in general. This brings a question for the future, which is specifying the optimal amount of useful information that could be used for a specific language.

The classifier that had the strongest correlation with FD was kNN ($\rho = -.7671$). From Table 4 we can see that the highest F1-scores were obtained for TOK+NER (F1=.605), LEM (F1=.637) and TOK (F1=.598), of which LEM be-ing the highest with its FD (.05928) being close to that of TOK+NER (.05891). In this case, assuming that classifier performance could be roughly predicted with FD, the maximum F1-score could be attainable with a dataset that has FD between .0593 (LEM), the current maximum, and .0702 (TOK). To confirm this assumption, we modified the datasets to reach FD from the above range. In particular, we removed punctuation of TOK+NER and LEM, which increased their FD to .0682 and .0684, respectively. With these datasets the performance of kNN classifier improved to F1=.6132 (TOK+NER, w/o punct.) and F1=.6205 (LEM, w/o punct.). The fact that these scores are higher than original TOK+NER (F1=.6050) and TOK (F1=.5976) supports the assumption that the score rises towards a maximum as we decrease the FD from .0702 (TOK) for a classifier that correlates negatively with FD. However, to fully confirm this claim, further experiments on larger datasets of different FD are required.

## 7 Conclusions

In this paper we presented our research on Feature Density and linguistically-backed embeddings, applied in cyberbullying detection. Both concepts are relatively novel to the field. We proposed a novel method of applying linguistically-backed embeddings in Convolutional Neural Networks and verified the correlation between the concept of Feature Density and classifier performance in the context of CB detection.

The proposed CNN model did not reach the highest performance, nonetheless, the classifier had on average higher performance than its competitors. The research also confirmed the effectiveness of SVM classifiers for CB detection also for English. Looking at the performance of CNN, word embeddings trained on plain tokens or lemmas achieved the highest scores for English, which suggests potential for improvement when a more robust embedding model is used. We also managed to confirm the assumption that there is a relation between classifier performance and FD, although the character and strength of this relation varies between classifiers.

In the near future we will explore further the potential of linguistically-backed embeddings and Feature Density. As this research focused on a cyberbullying dataset for English, in the near future we will verify the findings of this study on cyberbullying datasets in other languages, as well as for completely other classification tasks, to verify the extent to which the linguistically-backed embeddings can be improved, and how informative FD is for other languages and tasks.